\documentclass{article}
\usepackage{ijcai20}
\usepackage{times}
\usepackage{soul}
\usepackage{url}
\usepackage[hidelinks]{hyperref}
\usepackage[utf8]{inputenc}
\usepackage[small]{caption}
\usepackage{graphicx}
\usepackage{amsmath}
\usepackage{amsthm}
\usepackage{booktabs}
\usepackage{algorithm}
\usepackage{algorithmic}
\usepackage{amssymb}
\usepackage{bbm}

\title{Adversarial Attack via Dual-Stage Network Erosion}

\author{
Yexin Duan$^{1,2}$\and
Junhua Zou$^2$\and
Xingyu Zhou$^2$\and
Wu Zhang$^2$\and
Jin Zhang$^1$\And
Zhisong Pan$^2$\footnote{Corresponding author}\\
\affiliations
$^1$Zhenjiang Campus, Army Military Transportation University, Zhenjiang, China\\
$^2$Army Engineering University, Nanjing, China\\
\emails
\
}

\begin{document}

\maketitle

\begin{abstract}
Deep neural networks are vulnerable to adversarial examples, which can fool deep models by adding subtle perturbations. Although existing attacks have achieved promising results, it still leaves a long way to go for generating transferable adversarial examples under the black-box setting. To this end, this paper proposes to improve the transferability of adversarial examples, and applies dual-stage feature-level perturbations to an existing model to implicitly create a set of diverse models. Then these models are fused by the longitudinal ensemble during the iterations. The proposed method is termed Dual-Stage Network Erosion (DSNE). We conduct comprehensive experiments both on non-residual and residual networks, and obtain more transferable adversarial examples with the computational cost similar to the state-of-the-art method. In particular, for the residual networks, the transferability of the adversarial examples can be significantly improved by biasing the residual block information to the skip connections. Our work provides new insights into the architectural vulnerability of neural networks and presents new challenges to the robustness of neural networks.

\end{abstract}

\section{Introduction}
Deep neural networks (DNNs) have shown compelling accuracy in the field of visual tasks. However, it has been found that DNNs are vulnerable to adversarial examples, which are input examples perturbed by imperceptible perturbations, which are carefully crafted but can fool the networks into making wrong predictions ~\cite{szegedy2013intriguing,goodfellow2014explaining}.

The adversarial examples can be generated by white-box or black-box attacks. Since the internal information of the target model is usually not accessible, the black-box attacks remain a challenge. There are two main types of black-box methods, the query-based and the transfer-based. The query-based methods ~\cite{chen2017zoo,brendel2017decision} use queries to obtain the information of the target model so as to estimate the decision boundary, which makes the black-box attacks almost white-box attacks. However, they require a large number of queries, which would be impractical in real-world applications. It has been found that the adversarial examples can transfer, that is, the examples generated for one model under the white-box setting can successfully attack other unknown models ~\cite{szegedy2013intriguing,liu2016delving}. Hence, the transferability of adversarial examples can be leveraged to conduct black-box attacks.

Many techniques have been proposed to improve the transferability of adversarial examples, such as integrating the momentum term into the iterative process ~\cite{dong2018boosting}, applying random transformations to the input ~\cite{xie2019improving} and optimizing a perturbation over a set of translated images ~\cite{dong2019evading}. The standard model ensemble method ~\cite{liu2016delving,dong2018boosting} average the outputs ({\em e.g.}, logits) of multiple models to improve the adversarial attacks, which prevents adversarial examples from over-fitting to a specific model. These methods are either based on algorithm improvement, data augmentation or model input-output modification to improve the adversarial attacks, without considering the internal structural characteristics of the model. 

\begin{figure}[!t]
\small
\centering
\begin{minipage}{0.9\linewidth}
\centerline{\includegraphics[width=1\textwidth]{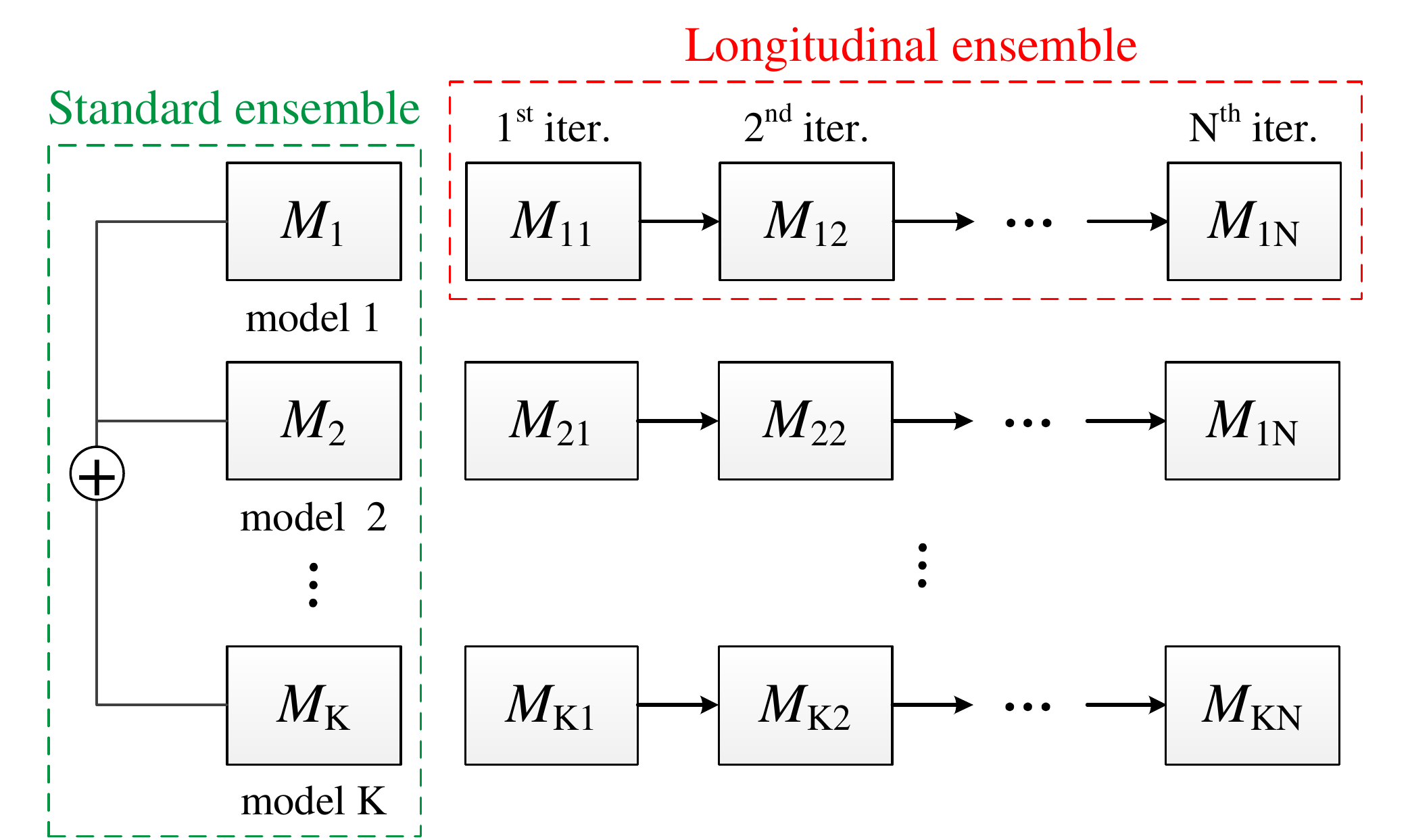}}
\end{minipage}
\caption{The illustration of standard ensemble and longitudinal ensemble.}
\label{fig:1}
\end{figure}

Recently, methods have been proposed to consider the model internal structures and parameters, such as Ghost Networks (GN) ~\cite{li2020learning}, which explores network parameter perturbations to potentially create a set of diverse models, and fuses these models by longitudinal ensemble. As illustrated in Fig. \ref{fig:1}, the standard ensemble requires averaging the outputs of different models. For the longitudinal ensemble, a set of diverse virtual models ({\em e.g.}, $\{ M_{11},M_{12},...,M_{\rm 1N} \}$) can be obtained from a base model ({\em e.g.}, $M_1$) by randomizing the perturbation during iterations of adversarial attack. GN improves the adversarial attacks and generates adversarial examples efficiently. However, the results in black-box attacks still leave a lot of room for improvement.

Motivated by the above discussion, in this paper, we propose a Dual-Stage Network Erosion (DSNE) method, which makes the network parameters more diversified to further improve the adversarial attacks. By imposing dual-stage erosion (feature-level perturbations) on the internal structures and parameters of the base networks on-the-fly, the forward and back propagation of the information flow would be modified, and multiple virtual models with similar decision boundaries are generated (``virtual" means that the generated models are not stored or trained). We call this operation ``model augmentation". Then these diversified virtual models are fused by the longitudinal ensemble during the iterations, which can alleviate the overfitting problem of iterative attacks, and the resultant adversarial examples are more likely to transfer across models.

Combining the proposed DSNE with any method ({\em e.g.}, momentum iterative method ~\cite{dong2018boosting}), we obtain more transferable adversarial examples with computation complexity similar to the baseline method. And the longitudinal ensemble can be easily combined with standard ensemble to further improve the transferability of adversarial examples. In addition, for the non-residual networks, more diversified virtual models are generated through the dual-stage network erosion, which enhances the effectiveness of transfer attacks. In particular, for the residual networks, since the classification performance improvement mainly comes from the skip connections, we adjust the role of skip connections in attacks. We find that the attack success rates significantly improved if the networks bias towards the skip connections. This indicates that the skip connections can expose more transferable information, which is beneficial for the adversarial examples to cross the decision boundaries.

In summary, our main contributions are as follows:

\begin{itemize}
\item{The proposed Dual-Stage Network Erosion (DSNE) method can generate more diverse virtual models and greatly improve the transferability of adversarial examples. }

\item{We find that the transferability of the resultant adversarial examples can be significantly enhanced by making the output of residual blocks of the residual network biased towards the skip connections.}

\item{We conduct extensive experiments both on normally trained models and robustly trained defense models, and the results demonstrate that our method can improve the black-box attacks with almost no extra computational cost.}

\item{The proposed dual-stage erosion method has wide compatibility, which can be imposed on both non-residual and residual networks, and can also be combined with different attack methods.}
\end{itemize}

\section{Related work}

Let $x$ be a clean input that can be correctly classified by a classifier $c(\cdot)$ as label $y$. An adversarial example $x^*$ can be obtain by adding imperceptible perturbations to $x$, which may fool the classifier, {\em i.e.}, $c(x^{*})\ne y$. For $L_{\infty}$ norm constraint, the allowed perturbation should be smaller than a threshold $\epsilon$ as $||x^{*}-x||_{\infty} \le \epsilon$. The attack objective is to maximize the cross-entropy loss function 
\begin{equation}
J(x^{*},y;\theta)=-\mathbbm{1}_y \cdot {\rm {log}({softmax}} (l(x^*))),
\label{eq1}
\end{equation}
where $\theta$ denotes the network parameters, $-\mathbbm{1}_y$ is the one-hot encoding of label $y$, and $l(x^*)$ is the classification logits of $x^*$, thus the adversarial deep learning problem can be expressed as 
\begin{equation}
\mathop{\rm argmax}\limits_{x^{*}}J(x^{*},y;\theta),\hspace{1em} {\rm s.t.} \hspace{0.5em} ||x^{*}-x||_\infty\leq\epsilon.
\label{eq2}
\end{equation}

\textbf{Iterative Fast Gradient Sign Method(I-FGSM).} I-FGSM ~\cite{kurakin2016adversariala} performs attack iteratively with a small step size. It initializes an adversarial example $x_{0}^{*}=x$ and the update equation is
\begin{equation}
x_{t + 1}^{*} = {\rm{Clip}}_{x}^\epsilon \{ x_t^{*} + \alpha  {\rm{sign}}({\nabla _{x_t^{*}}}J(x_t^{*},{y};\theta))\},
\end{equation}%
where $t$ is the $t$-th iteration and $\alpha$ is the step size. $\rm{sign(\cdot)}$ is the sign function. ${\rm{Clip}}_{x}^\epsilon \{ x'\}$  function performs per-pixel clipping of the image $x'$, it can be expressed as $ \min \left\{ {255,x + \epsilon ,\max \{ 0,x - \epsilon ,x'\} } \right\}$, so the result will be constrained within $\epsilon$-ball of the original image $x$. 

\textbf{Momentum Iterative Fast Gradient Sign Method (MI).} MI ~\cite{dong2018boosting} integrates the momentum term into I-FGSM to stabilize gradient update direction and avoid trapping into the local maximum, it can be expressed as
\begin{equation}
{g_{t + 1}} = \mu {g_t} + \frac{{{\nabla _{x_t^{*}}}J(x_t^{*},y;\theta)}}{{{{\left\| {{\nabla _{x_t^{*}}}J(x_t^{*},y;\theta)} \right\|}_1}}},
\end{equation}%
\begin{equation}
x_{t + 1}^{*} = {\rm{Clip}}_{x}^\epsilon {\rm{\{ }}x_t^{*} + \alpha  {\rm{sign(}}{g_{t + 1}}{\rm{)\} }},
\end{equation}%
where $g_{t}$ accumulates the iterated gradient vector of the loss function with a decay factor $\mu$.

\textbf{Translation-Invariant Method (TI).} TI ~\cite{dong2019evading} optimizes adversarial examples by convolving the gradient with a pre-defined kernel $W$, so that the generated adversarial examples would be less sensitive to the discriminative regions of the white-box model being attacked and have higher transferability. TI can be integrated into any gradient-based attack method, the integration of TI into the I-FGSM has the following update rule

\begin{equation}
 x_{t + 1}^{*} = {\rm{Clip}}_{x}^\epsilon \{ x_t^{*} + \alpha  {\rm{sign}}(W*{\nabla _{x_t^{*}}}J(x_t^{*},{y};\theta)\}.
\end{equation}%

\textbf{Ghost Networks (GN).} 
For non-residual networks, GN ~\cite{li2020learning} generates virtual networks by inserting the dropout layer densely to every block throughout the base network. Let $z_l$ be the activation in the $l$-th layer, $f_l$ be the function that satisfies $z_{l+1}=f_l(z_l)$ for the $l$-th and $(l+1)$-th layer, after applying dropout erosion, the output of $f_l$, {\em i.e.}, $g_l(z_l)$, is

\begin{equation}
{g_l}({z_l}) = {f_l}\left( {\frac{{{r_l}*{z_l}}}{{1 - {\Lambda}_b}}} \right),\;\;{r_l} \sim {\rm Bernoulli}(1 - {\Lambda}_b ),
\end{equation}%
where $*$ denotes an element-wise product and Bernoulli$(1 - {\Lambda}_b )$ means the Bernoulli distribution with the probability $p=(1 - {\Lambda}_b )$ of elements in $r_l$ being 1, {\em i.e.}, $p$ indicates the probability that $z_l$ is preserved. ${\Lambda}_b$ is defined as the magnitude of erosion, larger $\Lambda_b$ implies a heavier erosion on the source network, and vice versa.

For the networks with residual blocks, GN applies randomized modulating scalar $\lambda_l$ to the $l$-th residual block (see Fig. \ref{fig:3} (b)) by

\begin{equation}
{{{z}}_{l + 1}} = {\lambda _l}{z_l} + f_l({z_l},{w_l}),\;\;{\lambda _{l}} \sim {U}[1 - {\Lambda _u},1 + {\Lambda _u}],
\end{equation}%
where $\lambda_l$ is subject to uniform distribution, $z_l$ and $z_{l+1}$ are the input and output of the $l$-th residual block with the weights $w_l$, $f(\cdot)$ denotes the residual function. To keep the expected input of $z_l$ consisted after skip connection erosion, the mean of the uniform distribution is set to 1.

\section{Methodology}
GN explores network erosion to learn transferable adversarial examples, which can be applied both to single-model and multi-model attacks, and is compatible with various model structures and attack methods. However, there are several limitations: (1) GN generates a virtual model pool based on one-stage erosion to improve the transferability of adversarial examples, but the diversity of the network parameters is insufficient; (2) GN analyses the effect of erosion magnitude on classification accuracy, but does not analyse the effect on transferable attack performance, leading to inaccurate erosion magnitude and relatively low black-box attack success rates; (3) For ResNet-like networks, GN treats the skip connections (with an expected value of 1 for uniform distribution erosion) and residual modules equally. However, the main reason for the advanced performance of ResNet-like networks is the skip connections with the implementation of identity mapping, which can improve the information flow during forward and backward propagation, and enhance training efficiency and reduce test error ~\cite{srivastava2015highway,huang2016deep,veit2016residual}. Therefore, for the parallel structure of the skip connection and the residual module in a residual block, the skip connection should be made to transfer more information, so as to improve the transferability of adversarial examples.

To address these issues, firstly, the proposed DSNE method obtains more diversified networks by imposing dual-stage erosion on the base network, which further alleviates the overfitting phenomenon of iterative attack; secondly, DSNE optimizes the erosion magnitude for different networks according to the attack effect; thirdly, DSNE makes the output of each residual block biased towards the skip connection to mitigate the reduction of transferability information flow. 

In the following sections, we provide the detailed description of our DSNE method. The concept of model augmentation is proposed to introduce the principle of model diversification in Sec. \ref{sec 3.1}, then we introduce the dual-stage network erosion for non-residual and residual networks in Sec. \ref{sec 3.2} and Sec. \ref{sec 3.3}, respectively. The effect of erosion magnitude is analyzed in Sec. \ref{sec 4.2}, and comprehensive experiments are conducted for single-model and multi-model attacks in Sec. \ref{sec 4.3} and Sec. \ref{sec 4.4}, respectively.

\subsection{Model augmentation}
\label{sec 3.1}

Leveraging the transferability to attack is to generate adversarial examples under the white-box setting, and then use these examples to attack the unknown models. Traditional iterative attacks may easily overfit the parameters of the attacked white-box model, and thus making the generated adversarial examples rarely transfer to other models.

Different from the common methods, such as algorithm improvement ~\cite{dong2018boosting,dong2019evading}, data augmentation ~\cite{xie2019improving} and standard model ensemble ~\cite{liu2016delving,dong2018boosting,dong2019evading}, this paper alleviates the overfitting phenomenon by directly applying small parameter erosion ${{E}}( \cdot )$ to diversify the model, which satisfied $J(x,y;{{E}}(\theta)) \approx J(x,y;\theta)$ for any clean input $x$, by doing so, we derive a new model, and we call such derivation of models as { \textbf{model augmentation}}. Therefore, the constrained optimization problem in Eq. (\ref{eq2}) can be rewritten as 
\begin{equation}
\mathop{\rm argmax}\limits_{x^{*}}J(x^{*},y;{{E}}(\theta)),\hspace{1em} {\rm s.t.} \hspace{0.5em} ||x^{*}-x||_\infty\leq\epsilon.
\end{equation}

Due to the randomness of parameter erosion, each iteration will generate a new virtual model with similar decision boundaries, and then these multiple models generated at each iteration will be fused by the implicit longitudinal ensemble, making the resultant adversarial examples more transferable. The computation cost of the longitudinal ensemble attack is similar to that of base model iteration attack because network erosion requires little computation.

\subsection{Non-residual network erosion}
\label{sec 3.2}
For non-residual networks, to make the network parameters more diversified, the proposed DSNE method combines dropout and uniform distribution erosion, and the output of the $l$-th layer can be rewritten as

\begin{equation}
{g_l}({{{z}}_l}) = {f_l}\left( {\frac{{{r_l}*{\lambda _{{l}}}{z_l}}}{{1 - {\Lambda _b}}}} \right),
\end{equation}%
where $\lambda_l$ is drawn from the uniform distribution ${U}[1 - {\Lambda _u},1 + {\Lambda _u}]$, and $r_l$ is drawn from the Bernoulli distribution ${\rm Bernoulli}(1 - {\Lambda _b})$.

After applying the dual-stage erosion, the gradient of a loss function $J $ with respect to input $z_0$ in back-propagation from the $L$-th layer can be expressed as 
\begin{equation}
\frac{{\partial J }}{{\partial {z_0}}} = \frac{{\partial J }}{{\partial {{{z}}_L}}}\prod\limits_{{l} = 0}^L {\left( {\frac{{{r_l}}}{{1 - {\Lambda _b}}}*{\lambda _l}\frac{\partial }{{\partial {z_l}}}{f_l}\left( {\frac{{{r_l}*{\lambda _l}{z_l}}}{{1 - {\Lambda _b}}}} \right)} \right)}.
\label{eq_non-res_grad}
\end{equation}%

\subsection{Residual network erosion}
\label{sec 3.3}

Research ~\cite{srivastava2015highway,huang2016deep,veit2016residual} demonstrates that the identity mapping helps to learn to proceed in very deep networks, and there is some redundancy in the paths of the residual network, which shows that the random discard of some residual layers has little impact on the testing results. These techniques are mainly used to improve the training efficiency and testing accuracy of the residual networks. However, our work is to study attacking networks to improve the transferability of the adversarial examples. 

Residual networks ~\cite{he2016deep,he2016identity} are neural networks in which each layer consists of two subterms: an identity skip connection mapping and a residual module mapping. With $z_l$ as the input, the output of the $(l+1)$-th block is recursively defined as

\begin{equation}
z_{l+1} = z_l + f_l(z_l,w_l).
\end{equation}

Consider a 3-block residual network, from input $z_0$ to $z_3$, by expanding the recursion into the exponential number of nested items, we can make the structure of the residual network apparent, and obtain an unraveled view of the residual network ~\cite{veit2016residual}. Omitting the weights for clarity, the output can be expanded as 

\begin{equation}
\begin{aligned}
{z_3} =& {\rm{ }}{z_2}{\rm{ }} + {\rm{ }}{f_2}({z_2}) = [{z_1}{\rm{ }} + {f_1}({z_1})] + {\rm{ }}{f_2}({z_1}{\rm{ }} + {f_1}({z_1}))\\
 =& \left[ {{z_0} + {f_0}({z_0}){\rm{ }} + {f_1}({z_0} + {f_0}({z_0}))} \right] + {\rm{ }}{f_2} ( {z_0} + {f_0}({z_0}){\rm{ }} + \\
 & {f_1}({z_0} + {f_0}({z_0}))).
\end{aligned}
\label{eq_unraveled}
\end{equation}
  
\begin{figure}[!t]
\small
\centering
\begin{minipage}{0.9\linewidth}
\centerline{\includegraphics[width=1\textwidth]{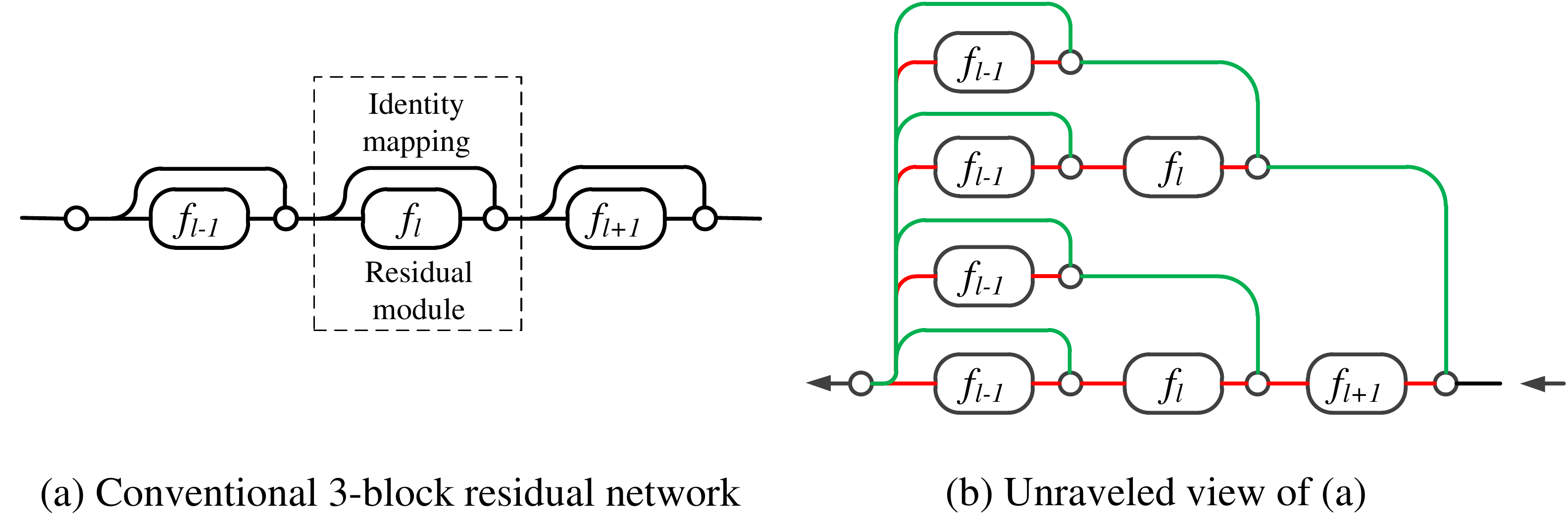}}
\end{minipage}
\caption{Conventional view (a) and unraveled view (b) of the residual network. Circular nodes denote junction point. The backpropagation paths of identity mapping and residual mapping are shown in green and red color, respectively.}
\label{fig:2}
\end{figure}
 
\begin{figure}[!t]
\small
\centering
\begin{minipage}{0.75\linewidth}
\centerline{\includegraphics[width=1\textwidth]{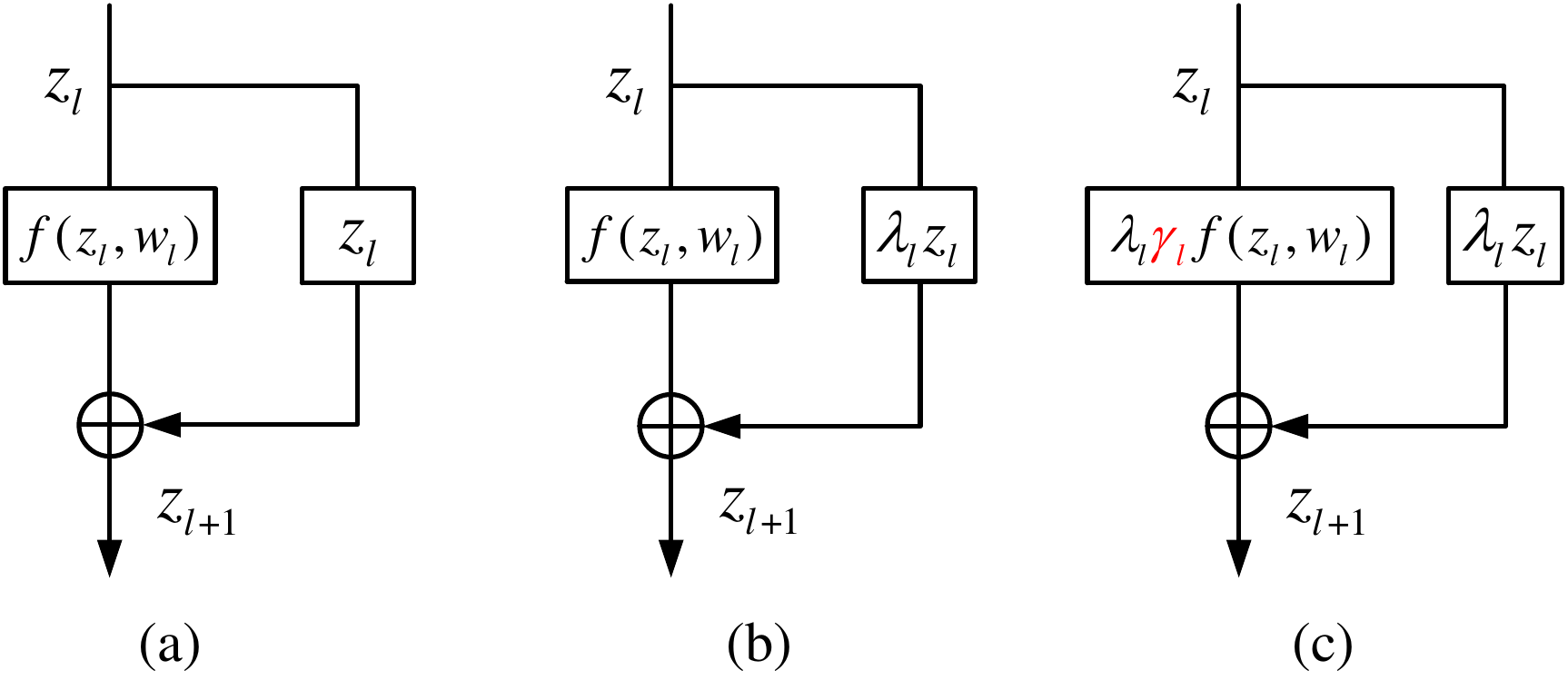}}
\end{minipage}
\caption{An illustration of (a) an original residual block, (b) the block after skip connection erosion and (c) the block after the dual-stage erosion.}
\label{fig:3}
\end{figure}

As shown in Fig. \ref{fig:2}, (a) is conventionally display form of the residual network, and (b) is the unraveled view as expressed in Eq. (\ref{eq_unraveled}). The reduction of residual gradients is accumulated along the backpropagation paths (red paths), while the identity mappings (green paths) facilitate the information propagation ~\cite{he2016identity}. Therefore, a bias toward identity mappings may expose more transferable information.

The network parameters are first learned by training the source network from scratch, then we apply dual-stage erosion on the identity mapping and the residual module in the $l$-th residual block (see Fig. \ref{fig:3} (c)) by 
\begin{equation}
{{{z}}_{l + 1}} = {\lambda _l}({z_l} + {\gamma _l}f({z_l},{w_l})),
\label{eq_res}
\end{equation}%
where $\lambda_l$ is drawn from the uniform distribution ${U}[1 - {\Lambda _u},1 + {\Lambda _u}]$, $\gamma_l$ is the bias factor and $0 < \gamma_l  \le 1$, such that the network is initially biased towards the shortcut connections which simply perform identity mapping. By doing so, it helps to improve the transferable information flow during forward and backward propagation, so as to enhance the attack effectiveness and obtain more transferable adversarial examples. It is worth noting that the model is not trained via Eq.(\ref{eq_res}).

The input of the $L$-th layer during inference can be written as

\begin{equation}
{{{z}}_L} = (\prod\limits_{l = 0}^{L - 1} {{\lambda _l}} ){z_0} + \sum\limits_{l = 0}^{L - 1} {(\prod\limits_{l = 0}^{L - 1} {\lambda _l} ){\gamma _l}f({z_l},{w_l})}.
\end{equation}
 
The gradient of a loss function $J$ with respect to input $z_0$ can be expressed as 
 
\begin{equation}
\frac{{\partial J }}{{\partial {z_0}}} = \frac{{\partial J }}{{\partial {z_L}}}\left( {(\prod\limits_{l = 0}^{L - 1} {{\lambda _l}} ) + \sum\limits_{l = 0}^{L - 1} {(\prod\limits_{l = 0}^{L - 1} {\lambda _l} ){\gamma _l}\frac{{\partial f({z_l},{w_l})}}{{\partial {z_0}}}} } \right).
\label{eq_res-grad}
 \end{equation}

The process of generating virtual models for non-residual or residual networks can be described in detail as follows: 1) conduct the uniform distribution erosion on the base network to obtain the perturbed network; 2) conduct the dropout or bias erosion on the perturbed network; 3) repeat step 1) and 2) to independently sample $\lambda$, $r$ or $\gamma$ for $N$ times ($N$ is the iteration number), and obtain a pool of virtual networks $M=\{ M_1,M_2,...,M_N \}$, which are fused by the implicitly longitudinal ensemble for attacks, {\em i.e.}, at the $i$-th iteration, it attacks the virtual model $M_i$ only.

Based on the above analysis, it can be inferred from the gradient of the loss function that a larger magnitude of erosion will have a greater influence on the source network, and deeper networks are influenced more easily according to the product rule. This is consistent with GN. 

DSNE is compatible with various attack methods, {\em e.g.}, combined with MI and TI, we get the TI-MI-DSNE attack, with $x_0^*=x$, it can be written as
\begin{equation}
{g_{t + 1}} = \mu {g_t} + \frac{W*\frac{{\partial J }}{{\partial {z_0}}}}{{{{\left\| W*\frac{{\partial J }}{{\partial {z_0}}} \right\|}_1}}},
\label{eq17}
\end{equation}%
\begin{equation}
x_{t + 1}^{*} = {\rm{Clip}}_{x}^\epsilon {\rm{\{ }}x_t^{*} + \alpha  {\rm{sign(}}{g_{t + 1}}{\rm{)\} }},
\label{eq18}
\end{equation}%
where $z_0=x_t^*$ is the input of the network at the $t$-th step, and $\frac{{\partial J }}{{\partial {z_0}}}$ for non-residual and residual networks are shown in Eq. (\ref{eq_non-res_grad}) and (\ref{eq_res-grad}), respectively. The TI-MI-DSNE combined with standard ensemble algorithm is summarized in Algorithm 1.

\begin{algorithm}[!htb]
\caption{TI-MI-DSNE combined with standard ensemble}
\label{alg:1}
\textbf{Input}: A clean example $x$ with label $y$; $K$ classifiers $c_{1},c_{2},...,c_{K}$; ensemble weights $w_{1},w_{2},...,w_{K}$;\\
\textbf{Parameter}: Perturbation size $\epsilon$; iteration number $N$ and momentum decay factor $\mu$; pre-defined kernel $W$; uniform distribution parameter  ${\Lambda _u} $, dropout parameter ${\Lambda _b}$ and scaling factor $\gamma$.\\
\textbf{Output}: An adversarial example $x^{*}$.

\begin{algorithmic}[1] 
\STATE $\alpha = \epsilon/N$;
\STATE $g_{0} = 0$; $x_0^* = x$;
\FOR {$t = 0$ to $ N-1 $}
\STATE Input $x_t^{*}$ and output the logits of $K$ classifiers: ${l_k}(x_t^{*}), k = 1,2,...,K$;
\STATE Fuse the logits as $ l(x_t^{*}) = \sum\nolimits_{k = 1}^K {{w_k}({l_k}(x_t^{*}))}$;
\STATE Get the cross-entropy loss $J$ based on $l(x_t^{*})$ and Eq. (\ref{eq1}) ;
\STATE Let $z_0=x_t^{*}$, for non-residual network and residual network, the gradient $\frac{{\partial J }}{{\partial {z_0}}}$ is calculated by Eq. (\ref{eq_non-res_grad}) and (\ref{eq_res-grad}), respectively;
\STATE Update the accumulated gradient $g_{t+1}$ and adversarial example $x_{t + 1}^{*}$ by Eq. (\ref{eq17}) and (\ref{eq18}), respectively;
\ENDFOR
\STATE \textbf{return}: $x^{*} = x_N^{*}$.
\end{algorithmic}
\end{algorithm}

\section{Experiments}

In this section, we evaluate our method by comparing the transfer attack success rates on the ImageNet dataset ~\cite{russakovsky2015imagenet} through a large number of experiments. We make our
codes public at \url{https://github.com/YeXinD/DSNE}.

\subsection{Experimental settings}
\textbf{Source Models.} We choose six models: Inception-v3 (Inc-v3) ~\cite{szegedy2016rethinking}, Inception-v4 (Inc-v4), Inception-ResNet-v2 (IncRes-v2) ~\cite{szegedy2017inception}, ResNet-v2-\{50, 101, 152\} (Res-\{50, 101, 152\}) ~\cite{he2016identity} as the source models.

\textbf{Target Models.} To evaluate the transferability
of the adversarial examples generated by the source
models, we consider fifteen target models, nine of which are normally trained models: Inc-v3, Inc-v4, IncRes-v2, Res-\{50, 101, 152\}, Densenet-169 (Dense-169) ~\cite{huang2017densely}, Xception-71 (Xcep-71) ~\cite{chollet2017xception}, and PNASnet-Large (PNAS) ~\cite{liu2018progressive}.  The other six are robustly trained defense models, including three ensemble adversarially trained models: Inc-v3$_{\rm ens3}$, Inc-v3$_{\rm ens4}$ and IncRes-v2$_{\rm ens}$~~\cite{tramer2017ensemble}, and the top-3 models in the NIPS 2017 Defense Competition: high-level representation guided denoiser (HGD)~~\cite{liao2018defense}, input transformation through random resizing and padding (R\&P)~~\cite{xie2017mitigating} and rank-3 solution\footnote{\url{https://github.com/anlthms/nips-2017/tree/master/mmd}} in the NIPS 2017 defense competition (NIPS-r3).

\textbf{Datasets.} It is less meaningful to study the attack success rates if the models cannot correctly classify the original images. Therefore, we randomly choose 5000 images from the ImageNet validation set, and these images are correctly classified by all source models. All these images are resized to $299 \times 299 \times 3$ beforehand.

\textbf{Baselines.} We mainly compare our DSNE method with MI ~\cite{dong2018boosting}, TI ~\cite{dong2019evading} and the corresponding GN ~\cite{li2020learning} methods. For all attack methods, the iteration number $N$ is set to 10, other hyper-parameters are set as in their original papers. We generate untargeted adversarial examples under maximum $L_\infty$ perturbation $\epsilon=16$ with respect to pixel values in $[0,255]$.

\subsection{Effect of erosion parameters}
\label{sec 4.2}
Due to the important influence of erosion parameters on the generation of strong transferable adversarial examples, a series of ablation experiments are conducted to study the effect of different erosion magnitude.

\begin{figure}[!t]
\small
\centering
\begin{minipage}{0.7\linewidth}
\centerline{\includegraphics[width=1\textwidth]{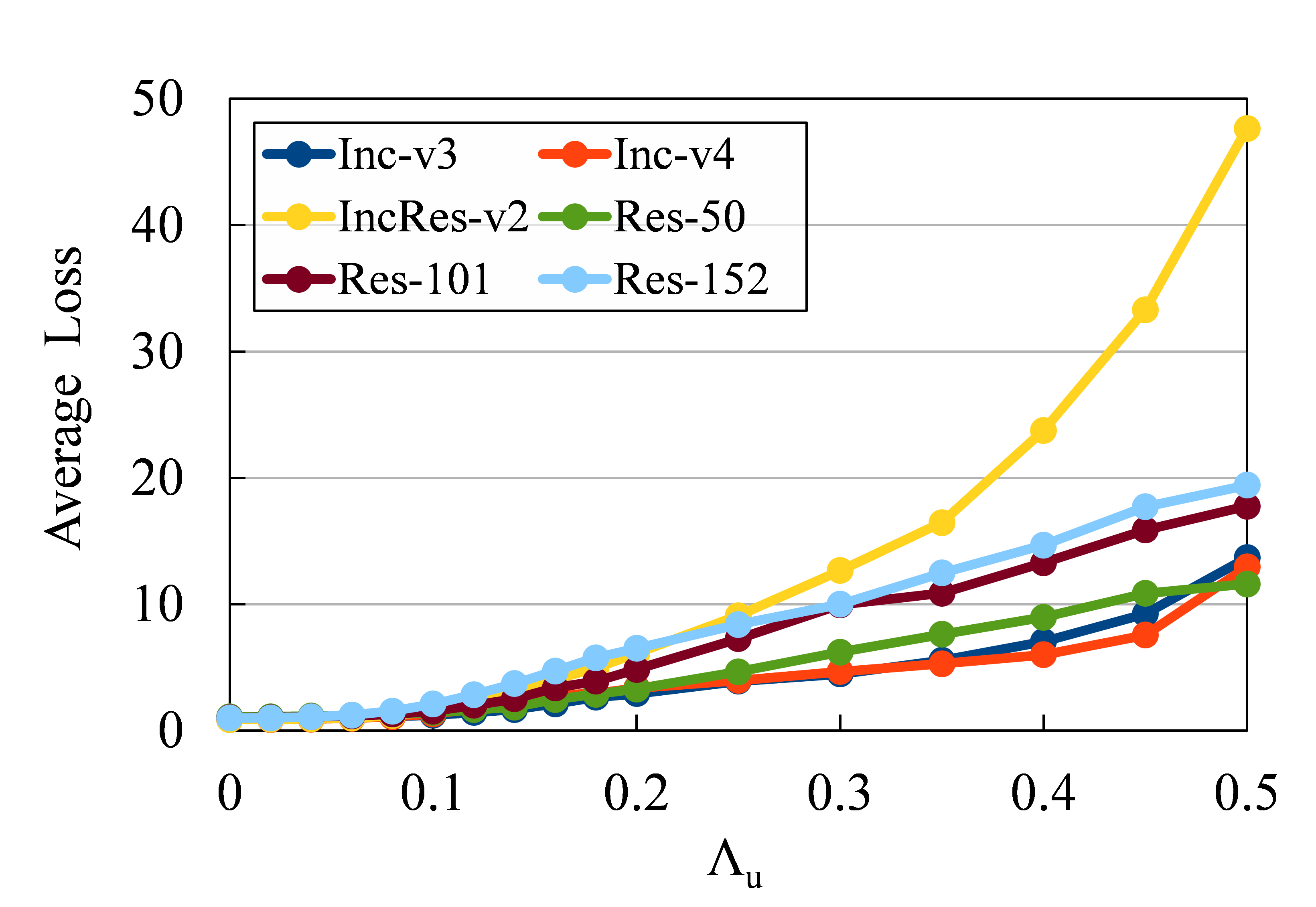}}
\end{minipage}
\caption{The average losses with different uniform distribution erosion magnitude of the six source models.}
\label{fig:4}
\end{figure}

\textbf{Uniform distribution parameter  ${\Lambda _u} $.} 
Uniform distribution parameter plays an important role in network diversity. We first verify the property of erosion parameter, {\em i.e.}, the effect of erosion on the classification performance of the model, with ${\Lambda _u} \in [0,0.5 ]$, where $\Lambda_u=0$ means no erosion on the source network. We input the clean images of the whole ILSVRC2012 validation set into the Inc-v3, Inc-v4, IncRes-v2, Res-50, Res-101 and Res-152, respectively. The average losses over all clean images for models with different erosion magnitude are shown in Fig. \ref{fig:4}.

\begin{figure}[!t]
\small
\centering
\begin{minipage}{0.93\linewidth}
\centerline{\includegraphics[width=1\textwidth]{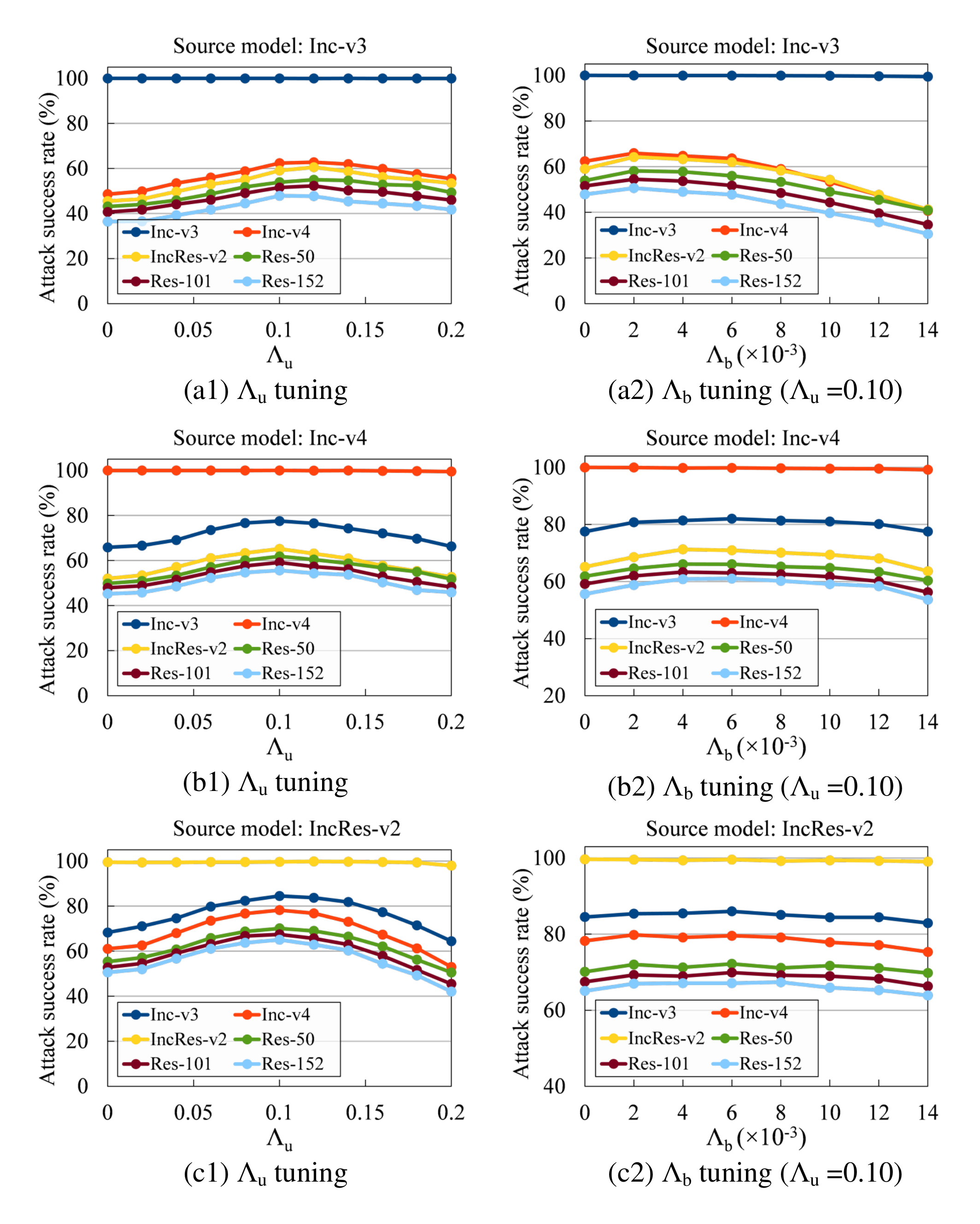}}
\end{minipage}
\caption{The attack success rates of Inception series networks with different erosion magnitude. The left column shows the $\Lambda_u$ tuning, the right column shows the $\Lambda_b$ tuning. }
\label{fig:5}
\end{figure}

\begin{figure}[!t]
\small
\centering
\begin{minipage}{0.915\linewidth}
\centerline{\includegraphics[width=1\textwidth]{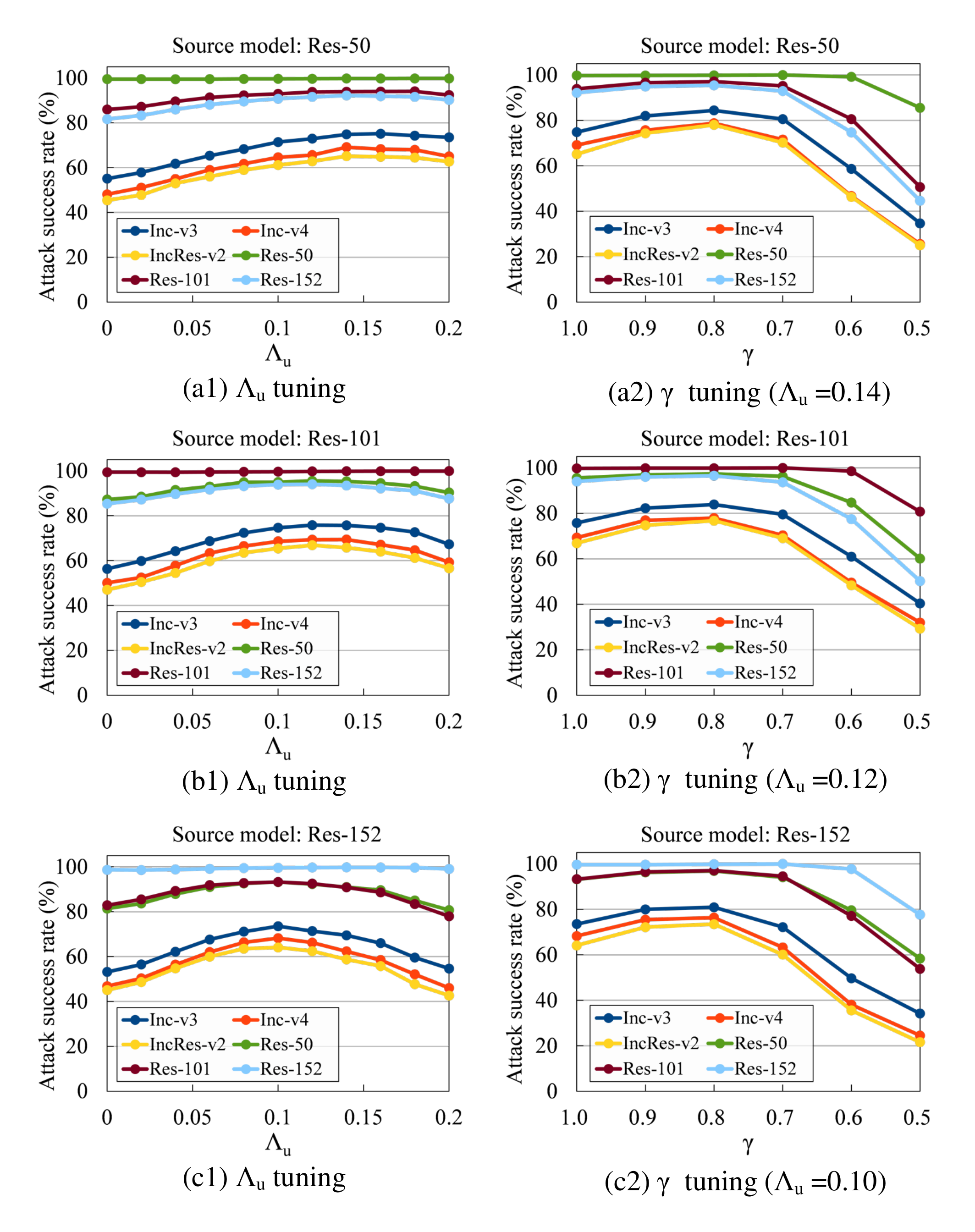}}
\end{minipage}
\caption{The attack success rates of ResNet series networks with different erosion magnitude. The left column shows the $\Lambda_u$ tuning, the right column shows the $\gamma$ tuning. }
\label{fig:6}
\end{figure}

It can be seen that with the increase of the erosion magnitude, the loss increases smoothly. Therefore, $J(x,y;{{E}}(\theta)) \approx J(x,y;\theta)$ is satisfied when this erosion magnitude is within a small range, which is consistent with the proposed concept of model augmentation in Sec. \ref{sec 3.1}. This rule also applies to the other two erosion parameters.

We then test the transferability with varying ${\Lambda _u} \in [0,0.2]$. The larger the $\Lambda_u$, the greater the erosion of the source network. The attack results of DSNE combined with MI method against six target models (one white-box and five black-box models) are illustrated in Fig. \ref{fig:5} (a1), (b1), (c1) and Fig. \ref{fig:6} (a1), (b1), (c1). It can be observed that the trends of attack success rates of all black-box attacks against different target models are consistent. Increasing the erosion magnitude $\Lambda_u$ tends to improve transferability until it exceeds a certain threshold. 

For the Inception series networks, all three source models have the highest attack success rates  when $\Lambda_u$ is set to 0.10; for the ResNet series networks, $\Lambda_u$ is set to 0.14 for ResNet-50, 0.12 for ResNet-101, and 0.10 for ResNet-152. It can be seen that for deeper networks, the erosion magnitude should be smaller, which is consistent with the previous inference that deeper networks are influenced more easily.

When the enhancement of transferable information brought by the network diversity is greater than the gradient information loss caused by network erosion, the attack success rates will increase. If the erosion magnitude is too large, the gradient information of the virtual networks will be quite different from that of the source network, and the obtained virtual network will not satisfy  $J(x,y;{E}(\theta)) \approx J(x,y;\theta)$, leading to the decrease of the attack success rates.

\textbf{Dropout parameter ${\Lambda _b}$.} For the Inception series networks, after tuning the uniform distribution parameter ${\Lambda _u}$, we test the transferability with varying dropout parameter ${\Lambda _b} \in [0,0.014]$, where ${\Lambda _b}=0$ means no dropout erosion on the network, and ${\Lambda _u}$ is set to 0.10. As shown in Fig. \ref{fig:5} (a2), (b2), (c2), the attack success rates increase until ${\Lambda _b}$ is greater than a certain value, 0.002 for Inc-v3, 0.004 for Inc-v4, 0.006 for IncRes-v2. The second stage erosion can make the virtual model more diverse, which further alleviates the overfitting problem and makes the resultant adversarial examples more transferable.

\textbf{Bias factor $\gamma$.} For residual networks, after tuning the erosion parameter  ${\Lambda _u}$, we investigate the effect of initial bias of the residual block towards identity mapping on transfer attack. We set the range of the bias factor $\gamma \in [0.5,1.0]$, where $\gamma =1.0$ means no bias in the residual blocks. 

Different layers of a neural network learn different levels of features, but the identity mapping can help preserve low-level features and avoid performance degradation when adding more layers, and allow unimpeded information flow across several layers ~\cite{srivastava2015highway,he2016identity}. While the reduction of residual gradients is accumulated along the backpropagation path, that is, the residual gradients at lower layers will be reduced more times than those at higher layers, the bias towards the identity mapping would help to preserve the low-level features (see Fig. \ref{fig:2} (b) and Eq. (\ref{eq_unraveled})) and expose more gradient information, so that the information flow bias towards the identity mapping (by reducing $\gamma$) could boost the adversarial attack and improve the transferability of adversarial examples.


As shown in Fig. \ref{fig:6} (a2), (b2), (c2), the trends of the influence of bias factor on transfer attack are consistent. And these three residual networks share the same optimal $\gamma$, {\em e.g.}, $\gamma=0.8$, which makes it easier to optimize the attack results. When the bias factor is too small, the class-relevant information will be excessively reduced, resulting in the failure of the model to obtain the correct class information and the useful gradient of the loss function, therefore, the attack success rates will decrease.

\begin{table*}[!htb]
\small
\centering
\caption{The attack success rates (\%) against the normally trained models. * indicates the white-box attacks. The adversarial examples are generated on each of the six source models, respectively. The best results are in bold.}
\resizebox{\textwidth}{38mm}{ 
\begin{tabular}{llcccccccccc}
\toprule[1.0pt]
Model                          & Attack                                                             & Inc-v3                                                       & Inc-v4                                                      & IncRes-v2                                                  & Res-50                                                     & Res-101                                                    & Res-152                                                    & Dense-169                                                  & Xcep-71                                                    & 
PNAS & Time(s)                                                       \\ \hline 
Inc-v3                         & \begin{tabular}[c]{@{}l@{}}MI\\ MI-GN\\ MI-DSNE\end{tabular} & \begin{tabular}[c]{@{}c@{}} \textbf{100.0}$^*$\\ 99.8$^*$\\ \textbf{100.0}$^*$\end{tabular} & \begin{tabular}[c]{@{}c@{}}48.6\\ 60.4\\ \textbf{65.9}\end{tabular}  & \begin{tabular}[c]{@{}c@{}}45.5\\ 59.1\\ \textbf{64.2}\end{tabular} & \begin{tabular}[c]{@{}c@{}}43.1\\ 53.5\\ \textbf{58.2}\end{tabular} & \begin{tabular}[c]{@{}c@{}}40.7\\ 49.7\\ \textbf{54.5}\end{tabular} & \begin{tabular}[c]{@{}c@{}}36.5\\ 45.5\\ \textbf{50.7}\end{tabular} & \begin{tabular}[c]{@{}c@{}}46.2\\ 56.0\\ \textbf{59.8}\end{tabular} & \begin{tabular}[c]{@{}c@{}}43.0\\ 54.7\\ \textbf{58.4}\end{tabular} & \begin{tabular}[c]{@{}c@{}}33.3\\ 41.5\\ \textbf{46.8}\end{tabular} & \begin{tabular}[c]{@{}c@{}}997.0\\ 1053.2\\ 1013.1\end{tabular}\\ \hline
Inc-v4                         & \begin{tabular}[c]{@{}l@{}}MI\\ MI-GN\\ MI-DSNE\end{tabular} & \begin{tabular}[c]{@{}c@{}}65.8\\ 79.6\\ \textbf{81.4}\end{tabular}   & \begin{tabular}[c]{@{}c@{}}\textbf{100.0}$^*$\\ 99.3$^*$\\ 99.8$^*$\end{tabular} & \begin{tabular}[c]{@{}c@{}}52.0\\ 67.5\\ \textbf{71.3}\end{tabular} & \begin{tabular}[c]{@{}c@{}}49.8\\ 63.2\\ \textbf{66.1}\end{tabular} & \begin{tabular}[c]{@{}c@{}}47.9\\ 61.4\\ \textbf{63.3}\end{tabular} & \begin{tabular}[c]{@{}c@{}}45.3\\ 58.6\\ \textbf{60.9}\end{tabular} & \begin{tabular}[c]{@{}c@{}}59.0\\ 71.7\\ \textbf{75.2}\end{tabular} & \begin{tabular}[c]{@{}c@{}}56.6\\ 72.4\\ \textbf{73.9}\end{tabular} & \begin{tabular}[c]{@{}c@{}}49.5\\ 64.9\\ \textbf{66.9}\end{tabular} & \begin{tabular}[c]{@{}c@{}}1424.6\\ 1665.2\\ 1710.3\end{tabular} \\ \hline
IncRes-v2 						& \begin{tabular}[c]{@{}l@{}}MI\\ MI-GN\\ MI-DSNE\end{tabular} & \begin{tabular}[c]{@{}c@{}}68.3\\ 79.6\\ \textbf{86.0}\end{tabular}   & \begin{tabular}[c]{@{}c@{}}61.1\\ 71.9\\ \textbf{79.6}\end{tabular}  & \begin{tabular}[c]{@{}c@{}}99.5$^*$\\ \textbf{99.7}$^*$\\ 99.6$^*$\end{tabular} & \begin{tabular}[c]{@{}c@{}}55.4\\ 64.8\\ \textbf{72.2}\end{tabular} & \begin{tabular}[c]{@{}c@{}}52.8\\ 61.8\\ \textbf{69.9}\end{tabular} & \begin{tabular}[c]{@{}c@{}}50.6\\ 59.3\\ \textbf{67.1}\end{tabular} & \begin{tabular}[c]{@{}c@{}}58.8\\ 68.1\\ \textbf{74.7}\end{tabular} & \begin{tabular}[c]{@{}c@{}}53.2\\ 62.3\\ \textbf{71.4}\end{tabular} & \begin{tabular}[c]{@{}c@{}}48.5\\ 57.4\\ \textbf{65.8}\end{tabular} & \begin{tabular}[c]{@{}c@{}}1548.5\\ 1744.8\\ 1833.1\end{tabular}\\ \hline
Res-50                         & \begin{tabular}[c]{@{}l@{}}MI\\ MI-GN\\ MI-DSNE\end{tabular} & \begin{tabular}[c]{@{}c@{}}55.1\\ 74.1\\ \textbf{84.4}\end{tabular} & \begin{tabular}[c]{@{}c@{}}48.1\\ 68.3\\ \textbf{78.8}\end{tabular}   & \begin{tabular}[c]{@{}c@{}}45.4\\ 64.9\\ \textbf{78.1}\end{tabular}  & \begin{tabular}[c]{@{}c@{}}99.5$^*$\\ 99.8$^*$\\ \textbf{99.9}$^*$\end{tabular} & \begin{tabular}[c]{@{}c@{}}85.9\\ 94.4\\ \textbf{97.1}\end{tabular} & \begin{tabular}[c]{@{}c@{}}81.7\\ 92.1\\ \textbf{95.5}\end{tabular} & \begin{tabular}[c]{@{}c@{}}60.2\\ 77.8\\ \textbf{86.1}\end{tabular} & \begin{tabular}[c]{@{}c@{}}48.8\\ 66.2\\ \textbf{77.5}\end{tabular} & \begin{tabular}[c]{@{}c@{}}41.9\\ 58.7\\ \textbf{70.8}\end{tabular} & \begin{tabular}[c]{@{}c@{}}922.5\\ 975.0\\ 975.1\end{tabular} \\ \hline
Res-101                        & \begin{tabular}[c]{@{}l@{}}MI\\ MI-GN\\ MI-DSNE\end{tabular}  & \begin{tabular}[c]{@{}c@{}}56.3\\ 75.5\\ \textbf{83.9}\end{tabular} & \begin{tabular}[c]{@{}c@{}}50.1\\ 69.1\\ \textbf{77.9}\end{tabular}   & \begin{tabular}[c]{@{}c@{}}47.0\\ 65.1\\ \textbf{76.8}\end{tabular}  & \begin{tabular}[c]{@{}c@{}}87.2\\ 95.6\\ \textbf{97.4}\end{tabular} & \begin{tabular}[c]{@{}c@{}}99.4$^*$\\ 99.8$^*$\\ \textbf{99.9$^*$}\end{tabular} & \begin{tabular}[c]{@{}c@{}}85.4\\ 93.9\\ \textbf{96.6}\end{tabular} & \begin{tabular}[c]{@{}c@{}}61.6\\ 79.4\\ \textbf{85.9}\end{tabular} & \begin{tabular}[c]{@{}c@{}}51.3\\ 69.3\\ \textbf{78.4}\end{tabular} & \begin{tabular}[c]{@{}c@{}}44.3\\ 60.1\\ \textbf{69.2}\end{tabular} & \begin{tabular}[c]{@{}c@{}}1241.1\\ 1319.5\\ 1385.7 \end{tabular} \\ \hline
Res-152                        & \begin{tabular}[c]{@{}l@{}}MI\\ MI-GN\\ MI-DSNE\end{tabular} & \begin{tabular}[c]{@{}c@{}} 53.2\\ 70.6\\ \textbf{80.9}\end{tabular} & \begin{tabular}[c]{@{}c@{}}46.8\\ 64.1\\ \textbf{76.3}\end{tabular}   & \begin{tabular}[c]{@{}c@{}}45.1\\ 60.0\\ \textbf{73.5}\end{tabular}  & \begin{tabular}[c]{@{}c@{}}81.4\\ 92.9\\ \textbf{96.9}\end{tabular} & \begin{tabular}[c]{@{}c@{}}82.9\\ 93.0\\ \textbf{97.1}\end{tabular} & \begin{tabular}[c]{@{}c@{}}98.7$^*$\\ 99.6$^*$\\ \textbf{99.8}$^*$\end{tabular} & \begin{tabular}[c]{@{}c@{}}58.5\\ 75.0\\ \textbf{83.9}\end{tabular} & \begin{tabular}[c]{@{}c@{}}48.7\\ 65.7\\ \textbf{76.5}\end{tabular} & \begin{tabular}[c]{@{}c@{}}42.3\\ 55.3\\ \textbf{67.0}\end{tabular}  & \begin{tabular}[c]{@{}c@{}}1615.0\\ 1688.1\\ 1812.4\end{tabular} \\
\bottomrule[1.0pt]
\end{tabular}}
\label{tab:1}
\end{table*}

\begin{table*}[!htb]
\small
\centering
\caption{The black-box attack success rates (\%) against the robustly trained defense models. The adversarial examples are generated on each of the six source models, respectively. The best results are in bold.}
\setlength{\tabcolsep}{3mm}{
\begin{tabular}{llccccccc}
\toprule[1.0pt]
Model     & Attack                                                                & Inc-v3$_{\rm ens3}$                                                 & Inc-v3$_{\rm ens4}$                                                & IncRes-v2$_{\rm ens}$ & HGD                                                        & R\&P                                                       & NIPS-r3  & Time(s)                                                  \\ \hline 
Inc-v3    & \begin{tabular}[c]{@{}l@{}}TI-MI\\ TI-MI-GN\\ TI-MI-DSNE\end{tabular} & \begin{tabular}[c]{@{}c@{}}30.3\\ 41.5\\ \textbf{43.7}\end{tabular} & \begin{tabular}[c]{@{}c@{}}28.1\\ 39.0\\ \textbf{41.2}\end{tabular} & \begin{tabular}[c]{@{}c@{}}20.0\\ 26.6\\ \textbf{30.5}\end{tabular} & \begin{tabular}[c]{@{}c@{}}20.3\\ 28.8\\ \textbf{31.4}\end{tabular} & \begin{tabular}[c]{@{}c@{}}17.0\\ 24.4\\ \textbf{27.4}\end{tabular} & \begin{tabular}[c]{@{}c@{}}21.2\\ 24.9\\ \textbf{30.3}\end{tabular} & \begin{tabular}[c]{@{}c@{}}1090.2\\ 1149.7\\ 1128.7 \end{tabular} \\ \hline
Inc-v4    & \begin{tabular}[c]{@{}l@{}}TI-MI\\ TI-MI-GN\\ TI-MI-DSNE\end{tabular} & \begin{tabular}[c]{@{}c@{}}32.3\\ 45.8\\ \textbf{46.2}\end{tabular} & \begin{tabular}[c]{@{}c@{}}31.3\\ 43.5\\ \textbf{44.7}\end{tabular} & \begin{tabular}[c]{@{}c@{}}23.5\\ 34.2\\ \textbf{34.6}\end{tabular} & \begin{tabular}[c]{@{}c@{}}24.2\\ \textbf{35.8}\\ 35.4\end{tabular} & \begin{tabular}[c]{@{}c@{}}21.6\\ \textbf{32.5}\\ \textbf{32.5}\end{tabular} & \begin{tabular}[c]{@{}c@{}}24.9\\ \textbf{37.6}\\ 37.2\end{tabular} & \begin{tabular}[c]{@{}c@{}}1520.2\\ 1759.6\\ 1767.4\end{tabular} \\ \hline
IncRes-v2 & \begin{tabular}[c]{@{}l@{}}TI-MI\\ TI-MI-GN\\ TI-MI-DSNE\end{tabular} & \begin{tabular}[c]{@{}c@{}}44.0\\ 53.4\\ \textbf{57.6}\end{tabular} & \begin{tabular}[c]{@{}c@{}}41.2\\ 49.9\\ \textbf{54.7}\end{tabular} & \begin{tabular}[c]{@{}c@{}}40.2\\ 48.7\\ \textbf{52.9}\end{tabular} & \begin{tabular}[c]{@{}c@{}}37.2\\ 46.1\\ \textbf{49.7}\end{tabular} & \begin{tabular}[c]{@{}c@{}}36.2\\ 43.7\\ \textbf{47.7}\end{tabular} & \begin{tabular}[c]{@{}c@{}}39.2\\ 48.5\\ \textbf{52.7}\end{tabular} & \begin{tabular}[c]{@{}c@{}}1626.5\\ 1779.2\\ 1919.4 \end{tabular} \\ \hline
Res-50    & \begin{tabular}[c]{@{}l@{}}TI-MI\\ TI-MI-GN\\ TI-MI-DSNE\end{tabular} & \begin{tabular}[c]{@{}c@{}}32.0\\ 47.2\\ \textbf{55.8}\end{tabular} & \begin{tabular}[c]{@{}c@{}}31.3\\ 45.0\\ \textbf{54.6}\end{tabular} & \begin{tabular}[c]{@{}c@{}}24.1\\ 36.9\\ \textbf{43.9}\end{tabular} & \begin{tabular}[c]{@{}c@{}}24.0\\ 37.0\\ \textbf{42.5}\end{tabular} & \begin{tabular}[c]{@{}c@{}}22.2\\ 33.7\\ \textbf{40.0}\end{tabular} & \begin{tabular}[c]{@{}c@{}}26.3\\ 40.0\\ \textbf{47.6}\end{tabular} & \begin{tabular}[c]{@{}c@{}}927.1\\ 1036.9\\ 1012.2\end{tabular} \\ \hline
Res-101   & \begin{tabular}[c]{@{}l@{}}TI-MI\\ TI-MI-GN\\ TI-MI-DSNE\end{tabular} & \begin{tabular}[c]{@{}c@{}}35.5\\ 48.0\\ \textbf{56.4}\end{tabular} & \begin{tabular}[c]{@{}c@{}}34.3\\ 46.2\\ \textbf{56.0}\end{tabular} & \begin{tabular}[c]{@{}c@{}}26.8\\ 37.4\\ \textbf{44.7}\end{tabular} & \begin{tabular}[c]{@{}c@{}}27.4\\ 38.1\\ \textbf{41.9}\end{tabular} & \begin{tabular}[c]{@{}c@{}}25.1\\ 35.1\\ \textbf{40.3}\end{tabular} & \begin{tabular}[c]{@{}c@{}}28.8\\ 41.1\\ \textbf{47.5}\end{tabular} & \begin{tabular}[c]{@{}c@{}}1269.9\\ 1352.7\\ 1495.0\end{tabular} \\ \hline
Res-152   & \begin{tabular}[c]{@{}l@{}}TI-MI\\ TI-MI-GN\\ TI-MI-DSNE\end{tabular} & \begin{tabular}[c]{@{}c@{}}34.7\\ 46.4\\ \textbf{55.5}\end{tabular} & \begin{tabular}[c]{@{}c@{}}33.8\\ 44.5\\ \textbf{55.4}\end{tabular} & \begin{tabular}[c]{@{}c@{}}27.5\\ 36.0\\ \textbf{45.4}\end{tabular} & \begin{tabular}[c]{@{}c@{}}27.1\\ 36.1\\ \textbf{42.9}\end{tabular} & \begin{tabular}[c]{@{}c@{}}25.4\\ 33.8\\ \textbf{41.7}\end{tabular} & \begin{tabular}[c]{@{}c@{}}29.5\\ 39.4\\ \textbf{48.5}\end{tabular} & \begin{tabular}[c]{@{}c@{}}1721.0\\ 1784.3\\ 1879.8\end{tabular} \\ \bottomrule[1.0pt]
\end{tabular}
\label{tab:2}}
\end{table*}

\subsection{Single-model attacks}
\label{sec 4.3}
In this section, we perform adversarial attacks on a single network. We craft adversarial examples on each of the six source models and test them on all fifteen target models. 

According to the discussion above, we select the optimized erosion parameters for each source model and combine our DSNE method with MI ~\cite{dong2018boosting} method to attack against the nine normally trained models, the comparison of the results are shown in Table \ref{tab:1}. Since TI ~\cite{dong2019evading} method is more effective for the defense models, we combine it to attack six robustly trained defense models, and the results are shown in Table \ref{tab:2}.

It can be seen that the black-box attack success rates of the proposed DSNE method are significantly higher than that of the baselines. Especially when the source model is the residual network, the average black-box attack success rates of our DSNE method is about 7\% ${\sim}$ 10\% higher than that of the Ghost Networks (GN) ~\cite{li2020learning} method. 

Note that the generated virtual networks are fused by the longitudinal ensemble, and these virtual models are not stored or trained, thus our attacks require similar time and space complexity to the baselines. 

In the last column of each table, we also list the running time as the computational cost of each attack method, each attack is run on an NVIDIA GTX 1080Ti GPU. It can be seen that our proposed DSNE method has similar computational costs to the baseline methods. 

\begin{figure}[!t]
\small
\centering
\begin{minipage}{0.8\linewidth}
\centerline{\includegraphics[width=1\textwidth]{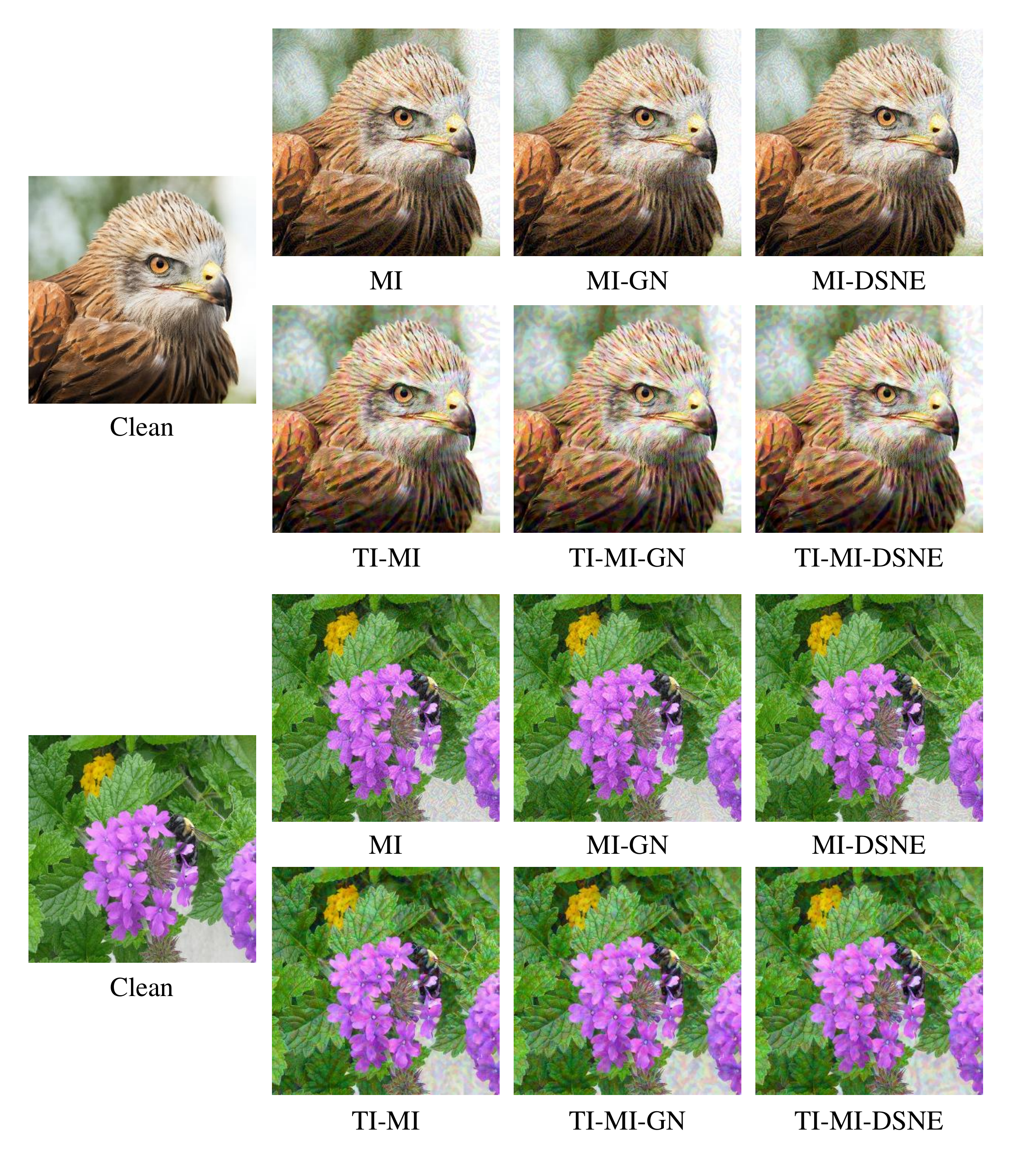}}
\end{minipage}
\caption{The adversarial examples generated by the proposed DSNE method and other baselines on the Inc-v3 model.}
\label{fig:8}
\end{figure}

\begin{table*}[!htb]
\centering
\small
\caption{The attack success rates (\%) against the normally trained models, and adversarial examples are generated on an ensemble of three source models. * indicates the white-box attacks. The best results are in bold.}
\resizebox{\textwidth}{27mm}{ 
\begin{tabular}{clcccccccccc}
\toprule[1.0pt]
Model                                                                       & Attack                                                                                       & Inc-v3                                                                           & Inc-v4                                                                          & IncRes-v2                                                                       & Res-50                                                                          & Res-101                                                                         & Res-152                                                                         & Dense-169                                                                       & Xcep-71                                                                         & PNAS                                                                           
& Time(s) \\ \hline 
\begin{tabular}[c]{@{}c@{}}Inc-v3\\ +\\ Inc-v4\\ +\\ IncRes-v2\end{tabular} & \begin{tabular}[c]{@{}l@{}}MI\\ MI-GN\\ MI-DSNE\\ TI-MI\\ TI-MI-GN\\ TI-MI-DSNE\end{tabular} & \begin{tabular}[c]{@{}c@{}}\textbf{100.0}$^*$\\ 99.8$^*$\\ 99.9$^*$\\ 99.6$^*$\\ 98.0$^*$\\ 98.9$^*$\end{tabular} & \begin{tabular}[c]{@{}c@{}}\textbf{99.7}$^*$\\ 98.5$^*$\\ 99.3$^*$\\ 97.7$^*$\\ 94.8$^*$\\ 96.7$^*$\end{tabular} & \begin{tabular}[c]{@{}c@{}}98.4$^*$\\ \textbf{99.3}$^*$\\ 98.9$^*$\\ 93.1$^*$\\ 94.8$^*$\\ 92.2$^*$\end{tabular} & \begin{tabular}[c]{@{}c@{}}74.8\\ 86.3\\ \textbf{87.1}\\ 62.7\\ 71.4\\ 72.1\end{tabular} & \begin{tabular}[c]{@{}c@{}}73.8\\ 85.5\\ \textbf{85.9}\\ 62.1\\ 70.3\\ 70.4\end{tabular} & \begin{tabular}[c]{@{}c@{}}72.1\\ 83.7\\ \textbf{84.7}\\ 61.0\\ 68.5\\ 68.6\end{tabular} & \begin{tabular}[c]{@{}c@{}}79.4\\ 89.1\\ \textbf{90.2}\\ 69.4\\ 78.8\\ 79.4\end{tabular} & \begin{tabular}[c]{@{}c@{}}77.9\\ 89.5\\ \textbf{90.1}\\ 66.5\\ 75.8\\ 76.5\end{tabular} & \begin{tabular}[c]{@{}c@{}}76.6\\ 86.9\\ \textbf{87.1}\\ 67.8\\ 74.4\\ 74.9\end{tabular} & \begin{tabular}[c]{@{}c@{}}3087.2\\ 3222.1\\ 3300.9\\ 3090.8\\ 3212.1\\ 3340.1\end{tabular} \\ \hline
\begin{tabular}[c]{@{}c@{}}Res-50\\ +\\ Res-101\\ +\\ Res-152\end{tabular}  & \begin{tabular}[c]{@{}l@{}}MI\\ MI-GN\\ MI-DSNE\\ TI-MI\\ TI-MI-GN\\ TI-MI-DSNE\end{tabular} & \begin{tabular}[c]{@{}c@{}}79.6\\ 91.6\\ \textbf{96.7}\\ 64.2\\ 74.3\\ 78.4\end{tabular}  & \begin{tabular}[c]{@{}c@{}}74.8\\ 89.3\\ \textbf{94.8}\\ 58.5\\ 68.8\\ 73.4\end{tabular} & \begin{tabular}[c]{@{}c@{}}73.9\\ 87.3\\ \textbf{94.8}\\ 57.3\\ 65.8\\ 70.8\end{tabular} & \begin{tabular}[c]{@{}c@{}}99.4$^*$\\ 99.7$^*$\\ \textbf{99.9}$^*$\\ 99.0$^*$\\ 98.2$^*$\\ 98.5$^*$\end{tabular} & \begin{tabular}[c]{@{}c@{}}99.4$^*$\\ 99.7$^*$\\ \textbf{99.9}$^*$\\ 99.0$^*$\\ 98.4$^*$\\ 98.2$^*$\end{tabular} & \begin{tabular}[c]{@{}c@{}}99.4$^*$\\ 99.7$^*$\\ \textbf{99.9}$^*$\\ 98.8$^*$\\ 98.3$^*$\\ 98.2$^*$\end{tabular} & \begin{tabular}[c]{@{}c@{}}81.7\\ 93.9\\ \textbf{97.3}\\ 63.8\\ 74.3\\ 77.1\end{tabular} & \begin{tabular}[c]{@{}c@{}}72.9\\ 88.5\\ \textbf{94.5}\\ 56.0\\ 66.5\\ 70.0\end{tabular} & \begin{tabular}[c]{@{}c@{}}72.7\\ 86.5\\ \textbf{92.1}\\ 59.9\\ 68.6\\ 71.6\end{tabular} & \begin{tabular}[c]{@{}c@{}}2707.0\\ 2836.3\\ 3189.3\\ 2859.8\\ 2972.2\\ 3281.4\end{tabular} \\ 
\bottomrule[1.0pt]
\end{tabular}}
\label{tab:3}
\end{table*}

\begin{table*}[!htb]
\small
\centering
\caption{The black-box attack success rates (\%) against the robustly trained defense models, and adversarial examples are generated on an ensemble of three source models. The best results are in bold.}
\setlength{\tabcolsep}{2mm}{
\begin{tabular}{clcccccc}
\toprule[1.0pt]
Model                                                                       & Attack                                                                                       & Inc-v3$_{\rm ens3}$  & Inc-v3$_{\rm ens4}$  & IncRes-v2$_{\rm ens}$  & HGD                                                                             & R\&P                                                                            & NIPS-r3                                                                         \\ \hline 
\begin{tabular}[c]{@{}c@{}}Inc-v3\\ +\\ Inc-v4\\ +\\ IncRes-v2\end{tabular} & \begin{tabular}[c]{@{}l@{}}MI\\ MI-GN\\ MI-DSNE\\ TI-MI\\ TI-MI-GN\\ TI-MI-DSNE\end{tabular} & \begin{tabular}[c]{@{}c@{}}35.3\\ 44.5\\ 44.9\\ 61.3\\ 70.0\\ \textbf{70.4}\end{tabular} & \begin{tabular}[c]{@{}c@{}}30.4\\ 39.1\\ 38.7\\ 59.1\\ 67.8\\ \textbf{68.5}\end{tabular} & \begin{tabular}[c]{@{}c@{}}18.6\\ 23.0\\ 23.0\\ 53.3\\ 62.1\\ \textbf{62.6}\end{tabular} & \begin{tabular}[c]{@{}c@{}}22.7\\ 23.6\\ 22.3\\ 56.5\\ 64.8\\ \textbf{64.9}\end{tabular} & \begin{tabular}[c]{@{}c@{}}18.5\\ 23.8\\ 23.5\\ 50.2\\ 60.1\\ \textbf{61.0}\end{tabular} & \begin{tabular}[c]{@{}c@{}}28.9\\ 37.6\\ 36.9\\ 54.6\\ 64.3\\ \textbf{64.7}\end{tabular} \\ \hline
\begin{tabular}[c]{@{}c@{}}Res-50\\ +\\ Res-101\\ +\\ Res-152\end{tabular}  & \begin{tabular}[c]{@{}l@{}}MI\\ MI-GN\\ MI-DSNE\\ TI-MI\\ TI-MI-GN\\ TI-MI-DSNE\end{tabular} & \begin{tabular}[c]{@{}c@{}}39.6\\ 51.6\\ 67.9\\ 57.4\\ 67.8\\ \textbf{76.0}\end{tabular} & \begin{tabular}[c]{@{}c@{}}33.9\\ 44.2\\ 61.2\\ 55.0\\ 65.7\\ \textbf{76.3}\end{tabular} & \begin{tabular}[c]{@{}c@{}}21.8\\ 28.4\\ 43.2\\ 47.6\\ 58.7\\ \textbf{66.7}\end{tabular} & \begin{tabular}[c]{@{}c@{}}33.7\\ 42.3\\ 49.2\\ 51.3\\ 60.6\\ \textbf{65.9}\end{tabular} & \begin{tabular}[c]{@{}c@{}}22.3\\ 28.9\\ 44.6\\ 46.1\\ 56.6\\ \textbf{63.7}\end{tabular} & \begin{tabular}[c]{@{}c@{}}32.2\\ 41.6\\ 59.2\\ 51.1\\ 62.0\\ \textbf{70.4}\end{tabular} \\ \bottomrule[1.0pt]
\end{tabular}}
\label{tab:4}
\end{table*}

We visualize two randomly selected clean images and their corresponding adversarial examples in Fig. \ref{fig:8}. All these adversarial examples are generated on Inc-v3 using different methods with the maximum perturbation $\epsilon=16$. Although the proposed DSNE method has significantly improved the black-box attack success rates, we can see that the magnitude of adversarial perturbations is almost the same as that of the baselines.

\subsection{Multi-model attacks}
\label{sec 4.4}

Research ~\cite{liu2016delving} demonstrated that attacking different models simultaneously can significantly improve the transferability of adversarial examples, which can also evaluate the robustness of the target models more accurately. We combine the standard ensemble and longitudinal ensemble, {\em i.e.}, the multi-model attack treats each longitudinal ensemble as a branch of the standard ensemble (seen in Fig. \ref{fig:1}). 

We attack the Inception series and ResNet series model ensembles, respectively. The success rates against nine normally trained models and six robustly trained models are summarized in Table \ref{tab:3} and Table \ref{tab:4}, respectively. Note that the TI method is originally used to attack robustly trained defense models, although here we use it to attack both normally trained and robustly trained models. It can be seen that similar to single-model attacks, our DSNE method can improve the transferability of the resultant adversarial examples significantly.

As shown in Table \ref{tab:3}, for the Inception series ensemble, the black-box attack performance of our DSNE method combined with MI is better than other methods. For the ResNet series ensemble, our DSNE method combined with MI consistently outperforms all other methods under both white-box and black-box settings. Compared with the strong baseline, {\em e.g.}, MI-GN, our MI-DSNE method improves the average black-box attack success rates by a large margin (about 6\%). Even only three source models are used, MI-DSNE achieves a high average black-box attack success rate (95.0\%), which verifies that the bias towards identity mapping makes the adversarial examples transfer more easily.

In Table \ref{tab:4}, for the Inception series ensemble, the DSNE method also shows superior attack performance. In addition, for the ResNet series ensemble, similar to the results of against normally trained models, DSNE combined with TI and MI consistently improves the transferability of the adversarial examples by a large margin, {\em e.g.} the average attack success rate is about 8\% higher than the TI-MI-GN. The results indicate that the structures of the deep networks are still vulnerable and the security of the networks can be enhanced from the structure design.

\section{Conclusion}
This paper studies enhancing the transferability of adversarial examples by eroding the internal parameters of the source network on-the-fly. First, we adopt the proposed dual-stage network erosion to augment the source models and make the models more diversified, which alleviates the overfitting problem of iterative attacks and makes the generated adversarial examples more transferable. Second, we fuse the generated virtual models by the longitudinal ensemble, which significantly enhances the black-box attack success rates with similar computational consumption. Particularly, for the residual network, we find that when the network is biased towards identity mapping, the transferability of the resultant adversarial examples will be improved significantly, the average attack success rates are about 6\% ${\sim}$ 10\% higher than that of the state-of-the-art method under the single-model and multi-model settings. Our work poses new challenges for the application of deep neural networks.

\bibliographystyle{named}
\bibliography{ijcai20}

\end{document}